%% file: main.tex
\documentclass[letterpaper, 10 pt, conference]{ieeeconf}  

\IEEEoverridecommandlockouts                              

\usepackage{graphics} 
\usepackage{epsfig} 
\usepackage{mathptmx} 
\usepackage{times} 
\usepackage{amsmath} 
\usepackage{amssymb}  
\usepackage{flushend}
\usepackage{hyperref}
\usepackage{gensymb}
\DeclareMathAlphabet{\mathcal}{OMS}{cmsy}{m}{n}

\DeclareMathOperator*{\argmax}{arg\,max}

\title{
Online SLAM with Any-time Self-calibration\\
and Automatic Change Detection
}

\author{Nima Keivan and Gabe Sibley*
\thanks{*This work was supported by Google, Toyota and MITRE}
\thanks{Nima Keivan and Gabe Sibley are with the Department of Computer Science, University of Colorado, Boulder, CO, USA
        {\tt\small nima.keivan|gsibley@colorado.edu}}%
}

\begin{document}

\maketitle
\thispagestyle{empty}
\pagestyle{empty}

\input{abstract}
\input{introduction}

\input{methodology}

\input{results}

\input{conclusions}

\bibliographystyle{plain}
\bibliography{references}

\end{document}

%% file: abstract.tex
\begin{abstract}
A framework for online simultaneous localization, mapping and self-calibration is presented which can detect and handle significant change in the calibration parameters. Estimates are computed in constant-time by factoring the problem and focusing on segments of the trajectory that are most informative for the purposes of calibration. A novel technique is presented to detect the probability that a significant change is present in the calibration parameters. The system is then able to re-calibrate. Maximum likelihood trajectory and map estimates are computed using an asynchronous and adaptive optimization. The system requires no prior information and is able to initialize without any special motions or routines, or in the case where observability over calibration parameters is delayed. The system is experimentally validated to calibrate camera intrinsic parameters for a nonlinear camera model on a monocular dataset featuring a significant zoom event partway through, and achieves high accuracy despite unknown initial calibration parameters. Self-calibration and re-calibration parameters are shown to closely match estimates computed using a calibration target.  The accuracy of the system is demonstrated with SLAM results that achieve sub-1\% distance-travel error even in the presence of significant re-calibration events.
\end{abstract}

%% file: introduction.tex
\section{Introduction} \label{sec:introduction}
Camera self-calibration is the inference of the intrinsic parameters of a camera without the explicit usage of a known calibration target. The motivation behind self-calibration in robotics is two-fold: it facilitates the use of computer vision for localization, mapping or scene understanding without requiring arduous calibration procedures, and also in the case of long-term autonomy, robustness is achieved against accidental changes in the calibration parameters. For the first case, a self-calibration methodology which continually estimates a single set of calibration parameters (such as camera intrinsics) will suffice. However, in order to deal with changes in the calibration parameters, the approach must facilitate the their re-estimation and detect the onset of the change event. Furthermore, it is desired to include other parameters such as extrinsics between different sensor modalities, time-delays or multiple cameras. For such a method to be useful in the case of long term autonomy however, it must be compatible with current developments in localization and mapping, as well as run in constant time to enable timely re-calibration in case of a change event.

With this motivation in mind, an approach is presented which enables the continuous estimation of camera intrinsics for a monocular setup in constant-time, while also simultaneously estimating the maximum likelihood camera and map parameters. The approach is based on a probabilistic method to detect when significant excitation of the calibration parameters is present in the motion to provide observability for estimation. It is able to deal with degenerate motions and non-linearities introduced due to unknown calibration parameters, which obviates the need for special initialization routines and motions. Probabilistic change detection indicates when the system should re-estimate parameters. Specific attention is paid to ensure past poses and landmarks are well estimated, even in the case of delayed observability of the calibration parameters. The approach is not exclusive to camera intrinsics and can be extended easily to estimate and detect changes in other calibration parameters, such as camera to IMU extrinsics, and time offsets. To the authors' knowledge, this is the first proposed solution to incorporate change detection for the estimation and re-estimation of calibration parameters in the SLAM setting.

\begin{figure}
\begin{centering}
\includegraphics[width=1.0\columnwidth]{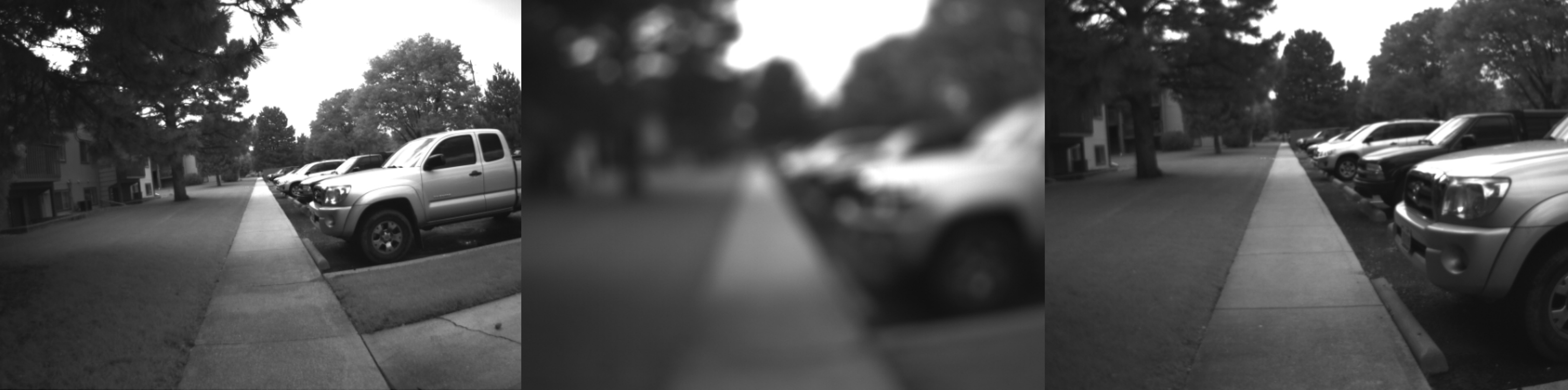}\caption{
Online self-calibration during a zoom event in a \~200m sequence. The left image was captured just before the zoom event. The center image was captured during the zoom, where focus is temporarily lost, causing a total loss of feature tracks. The right image shows an image captured just after the zoom. This event is automatically discovered and new accurate camera intrinsics are re-estimated online in real-time.
\label{fig:zoom_images}
}
\par\end{centering}
\end{figure}

\section{Related Work}  

The problem of self-calibration with varying camera intrinsics has received much attention in the literature in part due to the benefits outlined in section \ref{sec:introduction}. \cite{pollefeys1996euclidean} introduced a method to calibrate the varying focal length of a pinhole camera model in batch across images used for 3D reconstruction with all other intrinsic parameters known. \cite{heyden1997euclidean} presented a method based on bundle adjustment which optimized the focal length and principal point location parameters at each image location for a batch optimization over multiple images. Expanding upon this, \cite{Pollefeys99self-calibrationand} introduced a method to detect good portions of the trajectory for self-calibration. Both papers look at self-calibration in the batch setting. \cite{Agapito98self-calibrationof} presented a method which using the infinite homography constraint, estimates the focal length and principal point for a camera which only rotates, but does not translate. The rotation parameters along with the camera intrinsics are solved in batch using a nonlinear optimizer. \cite{Sturm99onplane-based} introduced a plane-based method for calibrating pinhole camera models in the batch setting. The principal point and focal lengths are estimated for each camera. \cite{Lourakis00cameraself-calibration} introduced a method for self-calibration of pinhole cameras using the SVD of the fundamental matrix between two views to derive constraints. These constraints are then solved in a nonlinear batch setting to obtain the focal length and principal point parameters of every camera. \cite{Frahm03cameracalibration} presented a method to calibrate the varying intrinsics of a pinhole camera in a batch setting, given the rotation of the camera was known. A solution was also offered to align the rotation sensor and camera data in time.

More recently, simultaneous solutions to the SLAM and self-calibration problem have been proposed, however to the author's best knowledge, all proposed online solutions assume constant calibration parameters. \cite{civera2009camera} proposed a method to recursively estimate 	camera and landmark 3D parameters as well as the intrinsic parameters of a nonlinear camera model in an online framework. To deal with the large non-linearities introduced by the unknown calibration parameters, a Sum of Gaussians (SOG) filter is used in lieu of an EKF. \cite{li2013high} presented a method based on the MSCKF \cite{mourikis2007multi} filter which also calibrates the IMU to camera extrinsics. \cite{martinelli2006automatic} introduced an EKF based method to estimate the calibration between an omnidirectional camera and robot odometry. \cite{Li2014} proposed a filtering solution based on the MSCEKF to estimate both the camera pose and also intrinsics and extrinsics for a non-linear camera model with rolling shutter and a commercial grade IMU in an online framework.

%% file: methodology.tex
\begin{figure*}
\begin{centering}
\includegraphics[width=2.0\columnwidth]{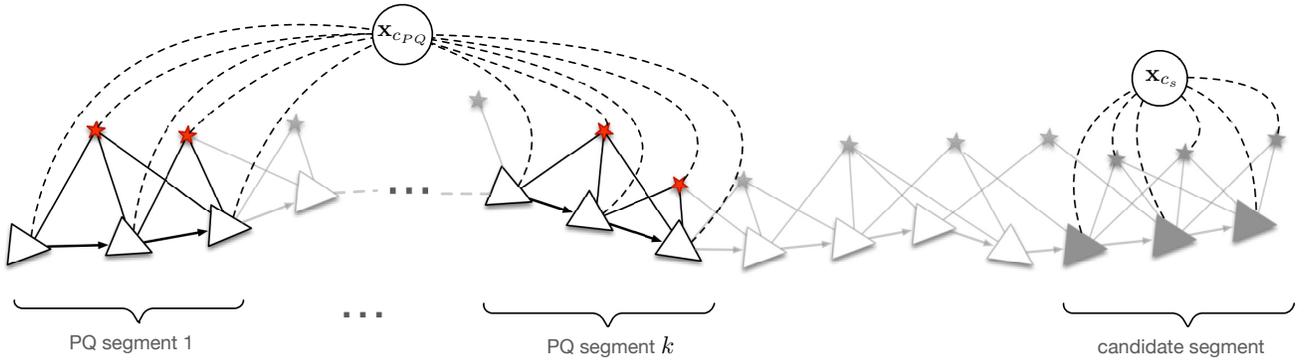}\caption{Graphical model showing the priority queue and candidate segments. Triangles represent camera pose parameters, while stars represent landmarks. Circles represent calibration parameters. In this figure, the priority queue segment size is 3 (refer to \cite{Keivan-ROBIO-14}). $\mathbf{x}_{c_{PQ}}$ refers to the calibration parameter estimate computed from the measurements in the priority queue, while $\mathbf{x}_{c_{s}}$ refers to the calibration parameter estimated computed solely from the measurements in the candidate segment. These two different estimates (and their posterior distributions) are used both to update the priority queue, and to decide if a change event has been detected.
\label{fig:pq_graphical_model}
}
\par\end{centering}
\end{figure*}

\section{Methodology}
The proposed method aims to continuously estimate the calibration parameters in constant-time, while also detecting the onset of a change event brought forth by perturbations to the sensors. Simultaneously, the maximum likelihood estimates of the camera pose and map parameters are desired. The required functionality can be composed as three sub-components: \textit{Constant Time Self-Calibration} is required in order to recursively estimate the maximum likelihood calibration parameters at any point in the trajectory, \textit{Change Detection} signals a high probability that the calibration has been perturbed during a change event, and  \textit{Adaptive SLAM estimation} is used to ensure maximum likelihood past and current camera and map parameters are estimated.

\subsection{Constant Time Self-Calibration} \label{sec:selfcal}
To recursively estimate the calibration parameters in constant-time, the approach described in \cite{Keivan-ROBIO-14} used. In order to aid the exposition of the overall self-calibration methodology, a brief summary of the method is presented here. The approach aims to obtain maximum likelihood values for the calibration parameters by selecting only the segments of the trajectory which provide the most information. In order to to assess this metric, a score is calculated based on the uncertainty of the calibration parameters as estimated by a particular \textit{candidate segment}. This score is then compared against the score of each segment stored in a fixed-size \textit{priority queue}. If the candidate segment score is better than the worst score in the priority queue, it is swapped into the priority queue. Once this update step takes place, a new estimate for the calibration parameters is obtained by using all segments in the priority queue jointly. As such, the priority queue will always contain the top $k$ most informative segments of the trajectory, where $k$ is a tuning parameter. The segments are of fixed size $m$, which is set as a constant tuning parameter. Figure \ref{fig:pq_graphical_model} shows the graphical model representing the priority queue, candidate segment and their respective estimates for the calibration parameters.

The joint probability distribution of the estimator state parameters given the measurements ($Z_{j}$) contained in the segment $j$ is given as 

\begin{align}
p\left(\mathbf{X}\vert Z_{j}\right)=p\left(\left\{ \mathbf{T}_{wc}\in SE3\right\} ,\left\{ \rho\right\} ,\mathbf{x}_{c}\vert Z_{j}\right)
\end{align}

\noindent where $\mathbf{T}_{wc}$ is the transformation from camera to world coordinates, $\rho$ is the landmark parameter given inverse depth parameterization \cite{Montiel-RSS-06} and $\mathbf{x}_{c}$ is the vector of calibration parameters. Note that the rotation component of the camera pose parameters $\left\{ \mathbf{R}_{wc}\in SO3\right\}$ is actually locally parameterized in the $so3$ tangent plane \cite{StrasdatPhd2012}. Using Bayes' Rule, the the joint probability can be factored as follows

\begin{align}
p\left(\mathbf{X},Z_{j}\right)=p\left(Z_{j}\vert \mathbf{X}\right)p(\mathbf{X})=\prod_{i=1}^{n}p\left(z_{i}\vert X\right)
\end{align}

here the likelihood term $p\left(Z_{j}\vert X\right)$ is factored due to the conditional independence assumption on individual measurements $z_i$, and the prior term $P(\mathbf{X})$ is omitted as the approach explicitly avoids a prior in favor of the priority queue. The optimal estimate for the parameter vector $\mathbf{X}$ is then one that would maximize the joint probability 

\begin{align}
\hat{\mathbf{X}}=\argmax_{\mathbf{X}} p\left(\mathbf{X},Z_{j}\right)
\end{align}

which would also be achieved by maximizing the likelihood term. Assuming a gaussian distribution over the parameter noise, the probability distribution over an individual term can be written as 

\begin{align}
p\left(z_{i}\vert\mathbf{X}\right)\propto\exp\left(-\frac{1}{2}\|z_{i}-h_i(\mathbf{X})\|_{\Sigma}^{2}\right)
\end{align}

\noindent where $\|\|_{\Sigma}^{2}$ denotes the squared mahalanobis distance, $z_{i}\in\mathbb{R}^{2}$ is the 2D pixel location of the landmark measurement and $h(\mathbf{X})\in\mathbb{R}^{2}$ is the measurement model, which predicts the 2D location of the measurement given the current state variables. The measurement model is defined as 

\begin{align}
h_{i}(\mathbf{X})=\mathcal{P}\left(\mathbf{p}_{r},\mathbf{X}\right)=\mathcal{P}\left(\mathbf{p}_{r},\mathbf{T}_{wc_{m}},\mathbf{T}_{wc_{r}},\rho_{l},\mathbf{x}_{c}\right) 
\end{align}

\noindent where $\mathcal{P}$ is a projection function which predicts the 2d pixel location of the projection of a given landmark into the measurement camera, given the 2d pixel location of the initial landmark observation ($\mathbf{p}_{2}$) in the reference camera, the transformation from the coordinates of the  reference and measurement cameras to the world coordinate frame ($\mathbf{T}_{wc_{r}}$ and $\mathbf{T}_{wc_{m}}$ respectively), the inverse depth of the landmark in the reference camera $\rho_l$, and the calibration parameter vector $\mathbf{x}_{c}$. In the current implementation, $\mathbf{x}_{c}$ consists of the 5 parameters of the FOV camera model \cite{Devernay01straightlines}:

\begin{align}
\mathbf{x}_{c}=\left[\begin{array}{ccccc}
f_{x} & f_{y} & c_{x} & c_{y} & w\end{array}\right]^T
\end{align}

\noindent where $f_{x}$, $f_{y}$, $c_{x}$, and $c_{y}$ are the $x$ and $y$ focal lengths and principal point coordinates respectively, and $w$ is a radial distortion parameter. Given this parameterization, the estimates for the camera poses $\hat{\mathbf{T}}_{wc_{n}}$, landmark inverse depths $\hat{\rho}_l$ and calibration parameters $\hat{\mathbf{x}}_{c}$ can be obtained via maximum likelihood estimation \cite{Triggs00bundleadjustment}. Furthermore, the normalized covariance matrix for the posterior distribution over the calibration parameters $\Sigma_{\mathbf{X_{c}}}^{\prime}$ given the measurements can be obtained by inverting the problem's Fisher information matrix $\mathcal{I}$ at convergence, and extracting the appropriate submatrix. This covariance matrix is normalized so as to remove the effects of the differing units (as described in \cite{Keivan-ROBIO-14}), and is then used to compute a score which is the priority queue update metric as previously described. Once an update operation takes place on the priority queue, all currently added segments will be used to jointly estimate a new value for the calibration parameters. The newly estimated parameters are then assigned to the frames in the set $\{n_{change},\ldots ,n_{current}\}$ where $n_{change}$ is the frame index of the last detected change event, and $n_{current}$ is the index of the current frame.

Due to the existence of critical motions which render some calibration parameters unobservable, the candidate segment covariance matrix is checked to ensure that it is full rank and well conditioned. If not, the candidate segment is discarded.

\subsection{Initialization} \label{sec:initialization}
Similar to the method described in \cite{Keivan-ROBIO-14}, a special initialization phase is used to bootstrap the priority queue and initialize the calibration parameters. A batch optimization is run over all state parameters (camera locations, parameter inverse depths and calibration parameters) from the most recent change index $n_{change}$ to the current frame $n_{current}$, much like the procedure used over a candidate segment. This joint estimation is run until the score \cite{Keivan-ROBIO-14} of the batch segment falls below a particular threshold. As the score is calculated from the entropy of the normalized posterior covariance, this threshold is a direct measure of the uncertainty of the posterior, and aims to prolong the batch optimization until the uncertainty over calibration parameters has been sufficiently reduced. 

Once this criteria is met, normal operation proceeds, where candidate segments are evaluated and added to the priority queue as necessary. 

\subsection{Change Detection} \label{sec:change_detection}

As shown in Figure \ref{fig:pq_graphical_model}, at each point in the trajectory, two posterior distribution estimates for the calibration parameters are available, each represented by a covariance matrix and mean. One is computed considering only the measurements within a candidate segment which is being evaluated for addition to the priority queue, with covariance  $\Sigma_s^{\prime}$, and another considering all measurements contained by the segments in the priority queue, with covariance $\Sigma_{PQ}^{\prime}$.

The priority queue posterior (with covariance $\Sigma_{PQ}^{\prime}$) represents the uncertainty over the calibration parameters considering the top $k$ segments in the trajectory. As these segments can have significant temporal separation, this distribution encodes the long term belief over the calibration parameters. Conversely, the candidate segment posterior (with covariance $\Sigma_s^{\prime}$) is calculated based on the most recent measurements and represents an instantaneous belief over the calibration parameters. 

A possible change in the actual calibration parameters would therefore manifest as a difference in the means represented by these two posterior distributions. The hypothesis test that two multivariate normal distributions with unknown and unequal covariance matrices have the same mean is known as the Multivariate Behrens-Fisher problem. The interested reader is directed to reviews for the univariate \cite{kim1998behrens} and multivariate \cite{park2009some} cases. Briefly, the null hypothesis of the test for the change detection case is as follows

\begin{align}
H_{0}=\mu_{PQ}=\mu_{s}
\end{align}

\noindent where $\mu_{PQ}$ is the mean estimated for the posterior distribution considering all the measurements in the priority queue, and $\mu_s$ is the mean estimated for the posterior distribution considering only the measurements in the candidate segment. For the purposes of this paper, the particular solution to the Behrens-Fisher problem used is the one proposed in \cite{yao1965approximate}. Given this method, the null hypothesis has an approximate F distribution which is given by 

\begin{align}
F_{p,v-p+1}\sim T^{2}\frac{v-p+1}{vp}
\end{align}

\noindent where the F distribution has degrees of freedom given by $p=\mathrm{dim}\left(\mathbf{x}_{c}\right)=5$ and $v-p+1$ with 

\begin{gather*}
v=\left[\frac{1}{n_{PQ}}\left(\frac{\mu_{d}^{T}\tilde{\Sigma}^{-1}\tilde{\Sigma}_{PQ}\tilde{\Sigma}^{-1}\mu_{d}}{\mu_{d}^{T}\tilde{\Sigma}^{-1}\mu_{d}}\right)^{2}+\frac{1}{n_{s}}\left(\frac{\mu_{d}^{T}\tilde{\Sigma}^{-1}\tilde{\Sigma}_{s}\tilde{\Sigma}^{-1}\mu_{d}}{\mu_{d}^{T}\tilde{\Sigma}^{-1}\mu_{d}}\right)^{2}\right]^{-1}\notag\\
\tilde{\Sigma}=\tilde{\Sigma}_{PQ}+\tilde{\Sigma}_{s}\notag\\
\tilde{\Sigma}_{PQ}=\frac{1}{n_{PQ}\left(n_{PQ}-1\right)}\Sigma_{PQ}\notag\\
\tilde{\Sigma}_{s}=\frac{1}{n_{s}\left(n_{s}-1\right)}\Sigma_{s}\notag\\
T^{2}=\mu_d^{T}\tilde{\Sigma}^{-1}\mu_d\notag
\end{gather*}

\noindent where $n_{PQ}$ and $n_s$ are the number of measurements in the priority queue and candidate segment respectively, and $\Delta\mu$ is the difference in the mean estimated by the priority queue and candidate segment given by $\mu_d=\mu_{PQ}-\mu_{s}$. Note that $\Sigma_{PQ}$ and $\Sigma_{s}$ are the un-normalized posterior covariances. Once the approximate F distribution is calculated, a corresponding p-value can be obtained, which can be compared to a level of significance parameter $\alpha$, and the null hypothesis is rejected if $p\leq\alpha$ where $\alpha=0.1$ is a tuning parameter which adjusts the sensitivity of the change detector. 

Using the previously outlined method for change detection, the uncertainty of both the priority queue and the candidate segment are considered in determining the probability of a change event. If the candidate segment mean is significantly different than the priority queue mean, but the entropy of its posterior is high relative to the priority queue posterior, it may have a lower probability of being a change event than a posterior distribution with a very low entropy relative to the priority queue, but significantly less deviation from the mean.

Sub-optimal tracking, motion blur and non-static features can all cause potentially misleading posterior distributions to be estimated for the candidate segment. As such, a simple heuristic is implemented to ensure that a single candidate segment which has a test p-value less than the level of significance parameter does not signal a change event. A number $n_{test}$ of candidate segments must consecutively reject the null hypothesis, where $n_{test}=3$ is a tuning parameter. If such an event does take place, the detected change frame index is set to $n_{change} = n_{current} - n_{test}$. This sets the starting index for a new set of calibration parameter assignments, as described in section \ref{sec:selfcal}.

Once a change event is detected, all of the current segments in the priority queue are removed, as they represent information contributing to a different set of calibration parameters, and the initialization routine outlined in section \ref{sec:initialization} is run once again to obtain an initial estimate over the new parameters. Once the initialization criteria has been met, the new priority queue is populate as per section \ref{sec:selfcal}.

\subsection{Adaptive SLAM}
The final component is an adaptive SLAM estimator which is able to incorporate the updates from the calibration estimation and and ensure that maximum likelihood current and past poses are re-estimated, in case calibration parameter estimates are updated. An immediate choice would be a recursive filtering method for SLAM estimation, \cite{leutenegger2013keyframe} \cite{li20133} \cite{mourikis2007multi}, However, with similar goals to that of the method in \cite{Keivan-ROBIO-14}, a linearized prior distribution is avoided in order to both keep the estimator consistent, and to enable the ease of integration of loop closure and relocalization constraints. Furthermore, as the aim of the system is to obviate the need for particular initialization motions (such as the SLAM wobble), the SLAM estimator must be able to handle potentially degenerate motions, where observability over calibration parameters does not eventuate until some time after initialization. In this case, past camera location and inverse depth parameters must be retroactively updated to ensure maximum likelihood map and trajectory estimates. Such non-linear updates (exacerbated by the non-linearities introduced \cite{civera2009camera} when intrinsic camera parameters change), motivate the use of an optimization based approach, which avoids marginalization altogether. Since consistency in filtering approaches is achieved by the use of \textit{first estimates} Jacobians \cite{li2013high}, a consistent filter would be further susceptible to non-linearities in the parameters and prior distribution. Considering the aforementioned points, the Adaptive Asynchronous Conditioning (AAC) \cite{Keivan-ISER-14} method is used to adaptively adjust the optimization window of an asynchronous bundle adjuster based on the error of the conditioning edge to the inactive part of the trajectory. 

When the calibration parameters are updated, the landmarks projecting in both the active and inactive portions of the trajectory will cause an increase in the conditioning error which will be lowered by expanding the optimization window and re-estimating past poses and landmarks. This is done asynchronously along with a constant-size windowed estimator which is conditioned on the inactive part of the trajectory and runs synchronously with the tracker.

The two AAC estimators run alongside the self-calibration estimation and are used to obtain the final estimates over camera position and landmark inverse depth parameters. This is also the case when the self-calibration estimator is in the initialization phase (section \ref{sec:initialization}). As the AAC estimator may enlarge the optimization window to span across a parameter change frame (indexed by $n_{change}$ described in section \ref{sec:change_detection}), care is taken to ensure that the appropriately assigned calibration parameters are used on either side of the change event.

Note that in section \ref{sec:results}, the AAC estimator is used with IMU information in order to maintain scale during an experimental trajectory, and also in order to carry the estimation through the loss of tracking caused by de-focus during the zoom operation (center image in Figure \ref{fig:zoom_images}). The formulation for the integration of IMU measurement is as per \cite{Keivan-ISER-14}. IMU measurements are solely used for scale-correct pose and map estimation. The self-calibration estimator does not use IMU information.

\subsection{Tracking and Keyframing} \label{sec:tracking}
To obtain feature correspondences between images, a method inspired by the tracking component of \cite{forster2014svo} is used, where the photometric error of a re-projected feature patch is directly minimized to obtain the new location of the feature. To initialize features, harris corners are used in a region of the image where not enough active tracks are present. Since the tracking method respects projective geometry, no RANSAC is required for frame-to-frame outlier rejection. Projective outliers however are still rejected in the bundle adjustment based on residuals. An NCC score is computed between the current and original feature patches and is thresholded (at 0.875) to reject feature patches which have changed too much in appearance.

For improved performance, and to solve the stationary camera problem, a keyframing \cite{MeiSCNR11} system is implemented based on heuristics on the rotation angle of the camera, traveled distance, and number of successful feature tracks from the last keyframe. If the criteria based on any of these metrics is met, a new keyframe is inserted and the current location of tracked features is used to add projections residuals. As such, if the camera is stationary or hardly moving, the features and consequently the pose of the camera are tracked with respect to the previous keyframe, but no new residuals are added to the problem.

%% file: results.tex
\begin{figure}
\begin{centering}
\includegraphics[width=1.0\columnwidth]{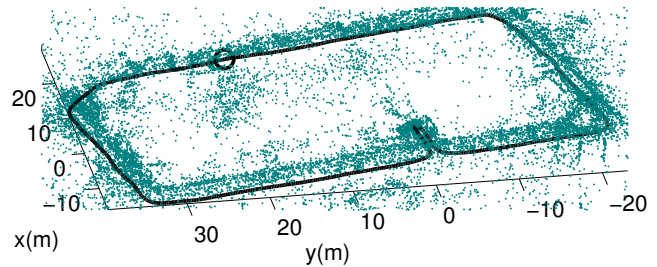}
\caption{Reconstruction and trajectory result from running the online self-calibration and AAC estimators on a 193.5m dataset captured on foot with the rig shown in Figure \ref{fig:rig}. The circle points to the location of the zoom event, where the focal length of the lens was manually adjusted from approximately 4mm to 8mm. The final translation error between the start and end poses is 0.8m or 0.42\% of the traveled distance. IMU measurements are used in the AAC estimator to ensure scale consistency in the map and trajectory estimates.}
\label{fig:ep_trajectory}
\par\end{centering}
\end{figure}	

\begin{figure}
\begin{centering}
\includegraphics[width=1.0\columnwidth]{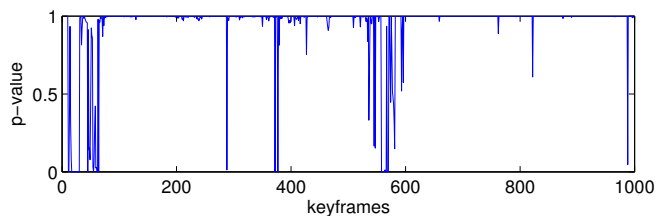}
\caption{p-values for the hypothesis test described in section \ref{sec:change_detection}, proposing that the priority queue and candidate segment window posterior means (corresponding to their best estimates of the calibration parameters) are different. p-values lower than the significance parameter $\alpha=0.1$ signal that calibration parameters have changed. The initial calibration starting at keyframe 0, as well as the zoom event at approximately keyframe 580 can be seen as numerous consecutive p-values smaller than the significance parameter.  }
\label{fig:ep_pvalue}
\par\end{centering}
\end{figure}	

\section{Results} \label{sec:results}
\begin{figure}
\begin{centering}
\includegraphics[width=1.0\columnwidth]{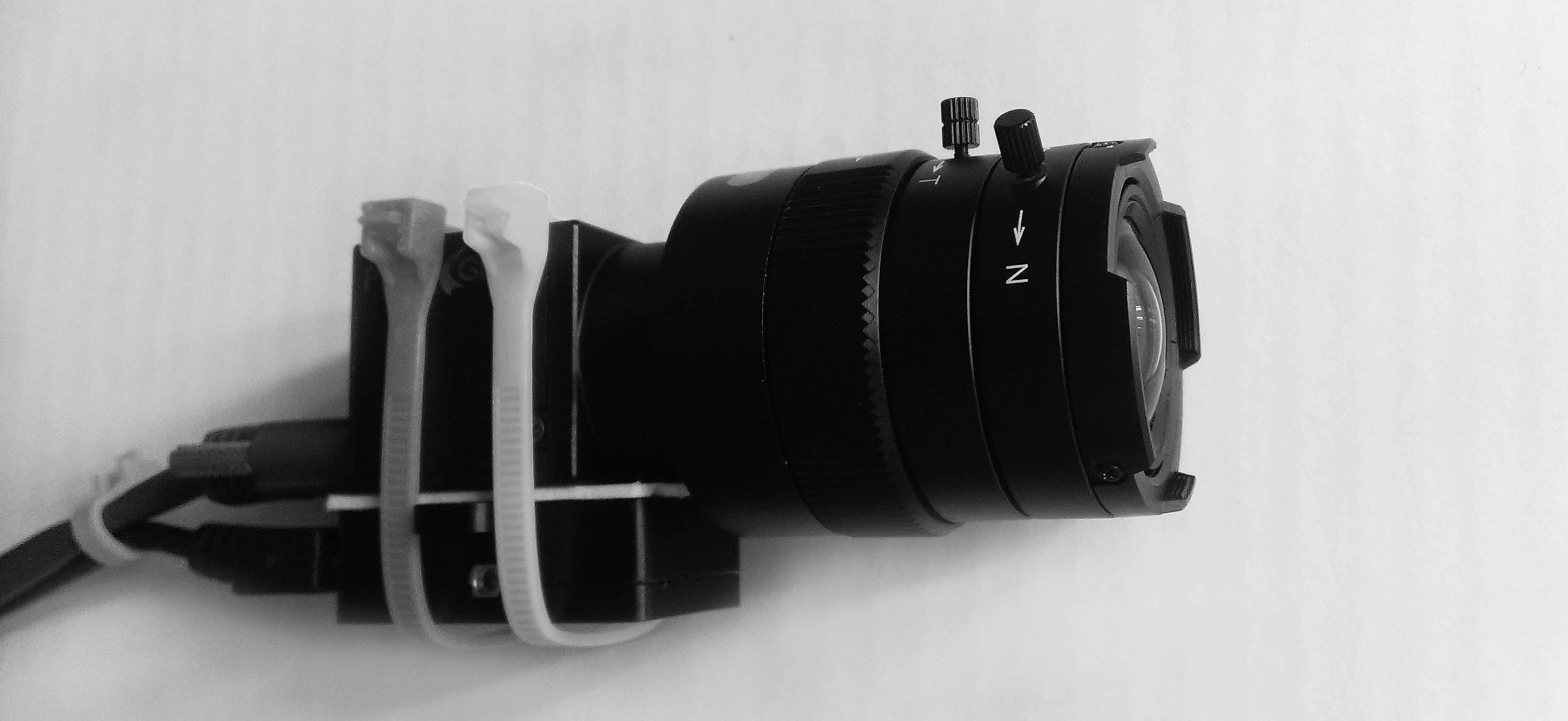}\caption{Experimental rig used to capture images (640x480 pixels at 30fps) and IMU data (at 200Hz). Consists of a USB camera attached to a USB commercial-grade IMU. The varifocal lens has adjustments for zoom and focus, with focal length variable between 4mm and 8mm.
\label{fig:rig}
}
\par\end{centering}
\end{figure}

The proposed method was run on real images captured from the visual-inertial rig in Figure \ref{fig:rig}. The rig consist of a grayscale global shutter camera recording at 640x480 pixels resolution and 30 frames per second, and a commercial IMU collecting accelerometer and gyroscope data at 200Hz. A varifocal lens is used with the camera, to allow the zoom and focus to be manually changed while collecting data. The images were used as described in section \ref{sec:tracking} to obtain feature tracks. The tracks were then made available to the SLAM and self-calibration estimators via a shared map structure. IMU information was additionally made available to the AAC estimator for scale consistency. In all experiments, the initial intrinsics estimates for the focal lengths were set to represent a field of view of $90\degree$, and the principal point initialized to be the image center. The distortion parameter $w$ was initialized to 1.

The first experiment was performed on an outdoor dataset captured by walking around a \~200m loop. A manual zoom event was performed (marked by the circle in the Figure) which consisted of changing the focal length of the camera from approximately 4mm to 8mm, as per the lens specifications. Due to the manual focus during the zoom, significant blur is introduced (as shown in Figure \ref{fig:zoom_images} which causes all feature tracks to be lost. In order to verify the accuracy of the reconstruction, IMU information is used in the AAC estimator (as per \cite{Keivan-ISER-14}) to both maintain a consistent scale through the trajectory, and also to carry the estimation through the segments where tracking loss is encountered. Scale loss is also encountered when the zoom event is introduced, as the intrinsics are not instantly known. The AAC estimator is used once again to re-acquire scale and maintain a maximum likelihood map and camera pose estimate, although this does not happen immediately. As the intrinsics are estimated by the self-calibration estimator, the AAC estimator simultaneously attempts to find the maximum likelihood scale, camera pose and map structure given the intrinsics. 
In order to properly utilize IMU measurements. the IMU to Camera transformation and time delay were calibrated offline in a batch optimization using a known target and set to constant during the experiment. Note that IMU information was not used in the self-calibration estimator, and the intrinsics were estimated solely from visual feature tracks. 

Apart from the heuristics outlined previously, the number of segments in the priority queue is set to 5 segments of size 10 keyframes each (for a description of these heuristics, the reader is directed to \cite{Keivan-ROBIO-14}). The tracker was configured to attempt to maintain 128 active feature tracks at any time. The keyframing heuristics were configured to add a new keyframe if the camera motion exceeds 0.1 radians in rotation, 0.2m in translation, or more than 20\% of tracks from the previous keyframe are lost.

The dataset features no special initialization motion, and involves simple forward walking from the first frame. The self-calibration and AAC estimators initialize the intrinsics and ensure maximum likelihood map and pose estimates, with consistent scale as obtained from the additional IMU measurements. Figure \ref{fig:ep_trajectory} shows the results of running the system on the dataset. The final translation error between the start and end camera positions is 0.8m, equating to a 0.42\% error per unit distance traveled.

Figure \ref{fig:ep_pvalue} shows the p-value plot (defined in section \ref{sec:change_detection} from the same dataset. It can be seen that at the beginning and around keyframe 580 the p-value is very small, indicating a large probability that the means of the priority queue and candidate segment posterior distributions are different. In these two cases, the indication is indeed warranted, as when the batch mode for self-calibration is activated (as described in \ref{sec:initialization}), the means of the priority queue and candidate segment posteriors will often be quite different, as a proper estimate of the calibration parameters is not yet available in the priority queue. However in several parts of the trajectory, spurious dips in the p-value can be seen, sometimes below the significance level parameter $\alpha$. These dips are the motivation for the heuristic introduced in section \ref{sec:change_detection}. Since the spurious dips are usually for a single candidate segment only, they do not falsely trigger the change detection system.

\begin{figure}
\begin{centering}
\includegraphics[width=1.0\columnwidth]{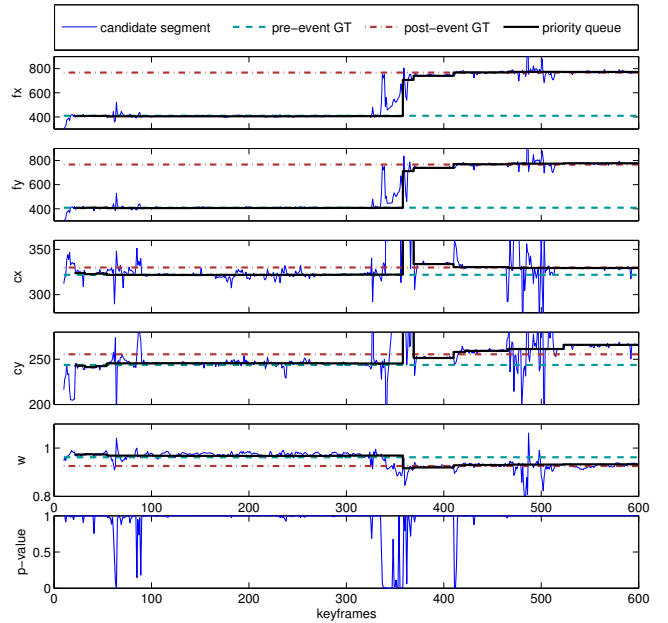}\caption{Comparisons between the ground truth intrinsics obtained from an offline calibration library \cite{calibu} and the priority queue and candidate segment estimates. The results are computed over a short outdoor dataset with simple forward walking motion. The plots are of the $x$ and $y$ focal lengths and principal point coordinates respectively, as well as the calibration parameter $w$ and the change detection p-value. The detection of the zoom event is clearly visible in the p-value plot.
}
\label{fig:cc_plots}
\par\end{centering}
\end{figure}	

The second experiment was undertaken with the aim of comparing the pre and post change event intrinsics with ground truth values obtained from an offline calibration method. The calibration library \cite{calibu} utilized uses a pre-defined target to estimate the intrinsics. Figure \ref{fig:cc_plots} shows the results of the ground truth values compared with the results obtained from running just the self-calibration estimator (no AAC) on images obtained from an outdoor dataset. Similar to the first experiment, the dataset consists of images captured during forward walking motion, with a manual zoom introduced partway through (detected approximately at the 400th keyframe). The figure shows the pre and post-zoom ground truth values, as well as the same parameter as estimated by the priority queue and the candidate segment. The priority queue estimate is shown as a stairstep plot, with steps occurring when a candidate segment is folded in, and a new estimate is computed.

It is notable that while at some points the candidate segment estimates vary wildly, the priority queue is not affected. These bouts of instability in the candidate segment can be caused by a variety of factors, including loss of tracking, motion blur, and outliers. The stability of the priority queue estimate during these portions is due to the high uncertainty in the posterior distribution of the parameters computed for the candidate segment. As such, these segments are not folded into the priority queue. 

After the zoom event, instability is also observed in the priority queue estimate, as all the segments in the queue are cleared, and the initialization routine is once again started (section \ref{sec:initialization}). This also explains the delay between the p-value change, and the first update to the priority queue. In this section, the batch optimization is run to initialize an estimate for the intrinsics. Once the initialization uncertainty reaches below a certain threshold (section \ref{sec:initialization}), the priority queue estimation is activated. For the most part, the estimates agree to a satisfactory level with the offline calibration values. Although theoretically the estimator should be consistent and unbiased, a more thorough examination of the results (such as with Monte-Carlo simulations) are in order, and will be slated for future work.

The p-value plot shows a similar trend to that of Figure \ref{fig:ep_pvalue}, where apart from a few spurious dips (which are ignored due to the heuristic introduced in section \ref{sec:change_detection}), the zoom event is clearly detected by a continuous dip in the p-value, suggesting a change event has occurred. 

\begin{figure}
\begin{centering}
\includegraphics[width=1.0\columnwidth]{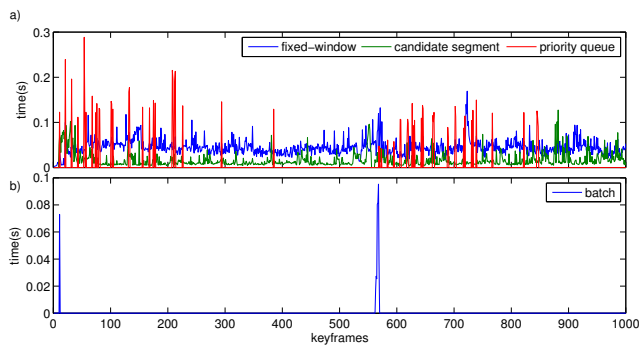}\caption{a) Time taken per keyframe for the fixed-window SLAM, candidate segment and priority queue estimators. It can be seen that the fixed-window and candidate segment estimators run in constant-time, while the priority queue estimator takes longer, but is only run when an update to the priority queue takes place and the calibration parameters need to be re-estimated.
b) Time taken for the batch optimization. It can be seen that the optimization is only invoked at the very beginning of the trajectory, and at the change event. Both times the optimization is run to initialize the calibration parameters as per section \ref{sec:initialization}.
}
\label{fig:timing}
\par\end{centering}
\end{figure}	

\subsection{Performance}
The current implementation runs at approximately 23fps on a Core i7 2.5Ghz processor laptop with a synchronous implementation of the AAC fixed-window and self-cal estimators. Figure \ref{fig:timing}a shows the timing information from the first experiment. It can be seen that the fixed-window estimator runs in constant-time during the entire trajectory. The estimator which obtains the posterior distribution over the candidate segment also runs in constant-time. The priority queue estimator is dormant for most of the trajectory, except when a candidate segment is swapped in, in which case a spike is observed as new intrinsics are estimated. Figure \ref{fig:timing}b shows the timings for the batch estimator. It can be seen that as stated during section \ref{sec:initialization}, the batch estimator is only run during initialization or when a change event is detected, in order to initialize the estimate for the calibration parameters. Currently, all estimators in Figure \ref{fig:timing} are run synchronously.  Run independently, the front-end tracker and fixed-window optimization achieves greater than 60fps.  A more optimized approach would be to run the front-end, candidate segment, and AAC solver asynchronously. This would both increase the average frame-rate of the system as well keep the front-end running at the fastest rate.

%% file: conclusions.tex
\section{Conclusions} \label{sec:conclusions}
This paper presents a method for online, constant-time self-calibration and automatic change detection and re-calibration. Experiments demonstrate that the system can estimate accurate calibration parameters, camera poses and landmark depths without any prior information. A filtering framework is explicitly avoided in favor of an adaptive asynchronous optimization \cite{Keivan-ISER-14} which provides the maximum likelihood current and past camera pose and landmark estimates. Rather than roll past information about the calibration parameters into a linearized prior distribution prone to inconsistency, a priority queue \cite{Keivan-ROBIO-14} is used to store the most-observable segments in the trajectory to estimate the calibration parameters. The approach enables "hands-off" initialization, where no specialized motion is needed. As parameters become observable over time the relevant segments are automatically included in the priority queue, and past camera pose and landmark parameters are updated when necessary.

A novel application of multivariate probabilistic change detection spurs the re-estimation of parameters if a significant change occurs. This causes re-calibration in situations where the physical sensor rig is perturbed.  To the authors' knowledge, this work is the first instance of online self-calibrating SLAM which can handle a significant change in the calibration parameters while still estimating the full maximum likelihood map and trajectory.

With additional IMU measurements to aid scale estimation, the system is able to achieve a distance-traveled error of 0.42\% even in the presence of a significant calibration change and total loss of tracking. Experiments show that parameter estimates from before and after a calibration-change event closely match values obtained via manual estimation with a calibration target.

While this paper has demonstrated real-time methods for two long-standing challenge-problems in robotics, namely 1) power-on-and-go self-calibration and 2) robust long-term SLAM in the face of model perturbation, the application of {\em probabilistic change detection} is perhaps the most compelling result; it is a powerful tool for introspection and model verification more generally.

%% file: main.bbl
\begin{thebibliography}{10}

\bibitem{Agapito98self-calibrationof}
L.~De Agapito, E.~Hayman, and I.~Reid.
\newblock Self-calibration of a rotating camera with varying intrinsic
  parameters.
\newblock In {\em In Proc 9th British Machine Vision Conf, Southampton}, pages
  105--114, 1998.

\bibitem{civera2009camera}
Javier Civera, Diana~R Bueno, Andrew~J Davison, and JMM Montiel.
\newblock Camera self-calibration for sequential bayesian structure from
  motion.
\newblock In {\em Proc. IEEE International Conference on Robotics and
  Automation (ICRA) 2009}, pages 403--408. IEEE, 2009.

\bibitem{Devernay01straightlines}
Frederic Devernay and Olivier Faugeras.
\newblock Straight lines have to be straight.
\newblock In {\em SPIE, volume 2567}, 2001.

\bibitem{forster2014svo}
Christian Forster, Matia Pizzoli, and Davide Scaramuzza.
\newblock Svo: Fast semi-direct monocular visual odometry.
\newblock In {\em Proc. IEEE Intl. Conf. on Robotics and Automation}, 2014.

\bibitem{heyden1997euclidean}
Anders Heyden and Kalle Astrom.
\newblock Euclidean reconstruction from image sequences with varying and
  unknown focal length and principal point.
\newblock In {\em Proc. IEEE Conference on Computer Vision and Pattern
  Recognition (CVPR) 1997}, pages 438--443. IEEE, 1997.

\bibitem{Keivan-ISER-14}
Nima Keivan and Gabe Sibley.
\newblock Asynchronous adaptive conditioning for visual-inertial slam.
\newblock In {\em Proceedings of the International Symposium on Experimental
  Robotics (ISER)}, 2014.

\bibitem{Keivan-ROBIO-14}
Nima Keivan and Gabe Sibley.
\newblock Constant-time monocular self-calibration.
\newblock In {\em Proceedings of the IEEE International Conference on Robotics
  and Biomimetics (ROBIO)}, 2014.

\bibitem{kim1998behrens}
Seock-Ho Kim and Allan~S Cohen.
\newblock On the behrens-fisher problem: A review.
\newblock {\em Journal of Educational and Behavioral Statistics},
  23(4):356--377, 1998.

\bibitem{leutenegger2013keyframe}
Stefan Leutenegger, Paul~Timothy Furgale, Vincent Rabaud, Margarita Chli, Kurt
  Konolige, and Roland Siegwart.
\newblock Keyframe-based visual-inertial slam using nonlinear optimization.
\newblock In {\em Robotics: Science and Systems}, 2013.

\bibitem{Li2014}
M.~Li, H.~Yu, X.~Zheng, and A.~I. Mourikis.
\newblock High-fidelity sensor modeling and calibration in vision-aided
  inertial navigation.
\newblock In {\em Proceedings of the IEEE International Conference on Robotics
  and Automation}, pages 409--416, Hong Kong, May 2014.

\bibitem{li20133}
Mingyang Li and Anastasios~I Mourikis.
\newblock 3-d motion estimation and online temporal calibration for camera-imu
  systems.
\newblock In {\em Proc. IEEE International Conference on Robotics and
  Automation (ICRA)}, 2013.

\bibitem{li2013high}
Mingyang Li and Anastasios~I Mourikis.
\newblock High-precision, consistent ekf-based visual--inertial odometry.
\newblock {\em The International Journal of Robotics Research}, 32(6):690--711,
  2013.

\bibitem{Lourakis00cameraself-calibration}
Manolis I.~A. Lourakis and Rachid Deriche.
\newblock Camera self-calibration using the kruppa equations and the svd of the
  fundamental matrix: The case of varying intrinsic parameters.
\newblock In {\em Research Report, INRIA SOPHIA-ANTIPOLIS}, 2000.

\bibitem{martinelli2006automatic}
Agostino Martinelli, Davide Scaramuzza, and Roland Siegwart.
\newblock Automatic self-calibration of a vision system during robot motion.
\newblock In {\em Robotics and Automation, 2006. ICRA 2006. Proceedings 2006
  IEEE International Conference on}, pages 43--48. IEEE, 2006.

\bibitem{MeiSCNR11}
Christopher Mei, Gabe Sibley, Mark Cummins, Paul~M. Newman, and Ian~D. Reid.
\newblock Rslam: A system for large-scale mapping in constant-time using
  stereo.
\newblock {\em International Journal of Computer Vision}, 94(2):198--214, 2011.

\bibitem{Frahm03cameracalibration}
Jan michael Frahm and Reinhard Koch.
\newblock Camera calibration with known rotation.
\newblock In {\em In Proceedings of IEEE Int. Conf. Computer Vision ICCV},
  pages 1418--1425, 2003.

\bibitem{Montiel-RSS-06}
J.~Montiel, J.~Civera, and A.~Davison.
\newblock Unified inverse depth parametrization for monocular slam.
\newblock In {\em Proceedings of Robotics: Science and Systems}, Philadelphia,
  USA, August 2006.

\bibitem{mourikis2007multi}
Anastasios~I Mourikis and Stergios~I Roumeliotis.
\newblock A multi-state constraint kalman filter for vision-aided inertial
  navigation.
\newblock In {\em Proc. IEEE International Conference on Robotics and
  Automation}, pages 3565--3572. IEEE, 2007.

\bibitem{park2009some}
Junyong Park and Bimal Sinha.
\newblock Some aspects of multivariate behrens-fisher problem.
\newblock {\em Bulletin of the Calcutta Statistical Association}, 61(241):125,
  2009.

\bibitem{Pollefeys99self-calibrationand}
Marc Pollefeys, Reinhard Koch, and Luc~Van Gool.
\newblock Self-calibration and metric reconstruction in spite of varying and
  unknown internal camera parameters.
\newblock In {\em International Journal of Computer Vision}, pages 7--25, 1999.

\bibitem{pollefeys1996euclidean}
Marc Pollefeys, Luc Van~Gool, and Marc Proesmans.
\newblock Euclidean 3d reconstruction from image sequences with variable focal
  lengths.
\newblock In {\em Computer Vision—ECCV'96}, pages 31--42. Springer, 1996.

\bibitem{calibu}
Autonomous Robotics and Perception Group.
\newblock Calibu camera calibration library.
\newblock \url{http://github.com/arpg/calibu}.
\newblock [Online; accessed September-2014].

\bibitem{StrasdatPhd2012}
Hauke Strasdat.
\newblock {\em {Local Accuracy and Global Consistency for Efficient Visual
  SLAM}}.
\newblock PhD thesis, Imperial College London, 2012.

\bibitem{Sturm99onplane-based}
Peter~F. Sturm and Stephen~J. Maybank.
\newblock On plane-based camera calibration: A general algorithm,
  singularities, applications.
\newblock In {\em Computer Vision and Pattern Recognition (CVPR)}, pages
  432--437, 1999.

\bibitem{Triggs00bundleadjustment}
Bill Triggs, Philip McLauchlan, Richard Hartley, and Andrew Fitzgibbon.
\newblock Bundle adjustment -- a modern synthesis.
\newblock In {\em Vision Algorithms: Theory and Practice, LNCS}, pages
  298--375. Springer Verlag, 2000.

\bibitem{yao1965approximate}
Ying Yao.
\newblock An approximate degrees of freedom solution to the multivariate
  behrens fisher problem.
\newblock {\em Biometrika}, pages 139--147, 1965.

\end{thebibliography}
